\pdfoutput=1

\documentclass[11pt]{article}

\usepackage[review]{acl}

\usepackage{times}
\usepackage{latexsym}
\usepackage{graphicx}
\usepackage{listings}

\usepackage[T1]{fontenc}

\usepackage[utf8]{inputenc}

\usepackage{hyperref}       
\usepackage{url}            
\usepackage{booktabs}       

\usepackage{amsfonts}       
\usepackage{nicefrac}       
\usepackage{microtype}      
\usepackage{xcolor}         
\usepackage{graphicx}
\usepackage{caption}
\usepackage{subfig}
\usepackage[ruled,vlined]{algorithm2e}
\usepackage{algorithmic}
\usepackage{multirow}
\usepackage{makecell}
\usepackage{multicol}
\usepackage{listings, xcolor}
\usepackage{enumitem}

\usepackage{setspace}
\usepackage[T1]{fontenc}

\usepackage[utf8]{inputenc}

\usepackage{microtype}

\usepackage{inconsolata}
\UseRawInputEncoding
\newcommand{\methodname}{EERPD}
\title{\methodname{}: Leveraging Emotion and Emotion Regulation for Improving Personality Detection}

\author{Zheng Li$^{1}$, Dawei Zhu$^{1}$, Qilong Ma$^{2}$, Weimin Xiong$^{1}$, Sujian Li$^{1}$ \\
$^1$State Key Laboratory for Multimedia Information Processing, \\
School of Computer Science, Peking University \\ 
$^2$School of Software, BNRist, Tsinghua University \\
\texttt{\{lycheelee, dwzhu,lisujian\}@pku.edu.cn}\\
\texttt{mql22@mails.tsinghua.edu.cn}\\
}

\begin{document}
{\makeatletter\acl@finalcopytrue
  \maketitle
}
\begin{abstract}

Personality is a fundamental construct in psychology, reflecting an individual's behavior, thinking, and emotional patterns. Previous researches have made some progress in personality detection, primarily by utilizing the whole text to predict personality. However, these studies generally tend to overlook psychological knowledge: they rarely apply the well-established correlations between emotion regulation and personality. Based on this, we propose a new personality detection method called \textbf{\methodname{}}. This method introduces the use of emotion regulation, a psychological concept highly correlated with personality, for personality prediction. By combining this feature with emotion features, it retrieves few-shot examples and provides process CoTs for inferring labels from text. This approach enhances the understanding of LLM for personality within text and improves the performance in personality detection. Experimental results demonstrate that \methodname{} significantly enhances the accuracy and robustness of personality detection, outperforming previous SOTA by 15.05/4.29 in average F1 on the two benchmark datasets.

\end{abstract}

\section{Introduction}

As a fundamental construct in psychology, \textit{personality} reveals the true nature of the individual and creates a certain impression on others~\citep{jung1959archetypes, corr2009cambridge, jung1959archetypes}. 
With the advancement of Natural Language Processing (NLP) technologies, there has been an growing interest in automatic detection of personality~\citep{petrides2018theory,yang2023psycot}, which 
plays a pivot role in numerous human-oriented NLP applications, such as psychological health assessment \cite{wilkinson2001attachment}, personalized recommendation systems \cite{hu2010study}, and human-computer interaction \cite{pocius1991personality}.

\begin{figure}[t]
\centering
\includegraphics[width=\linewidth]{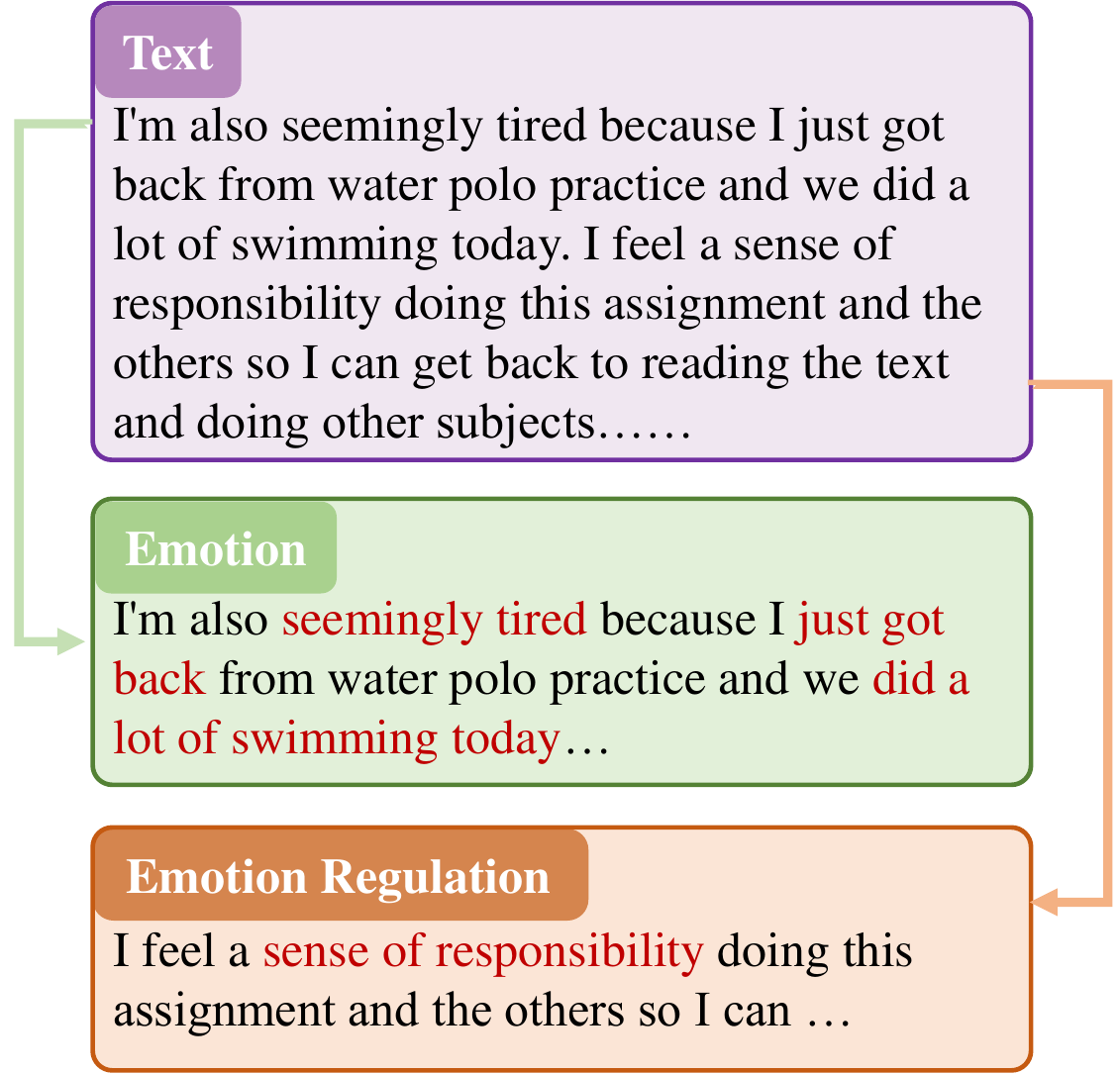}
\caption{Examples for emotion and emotion regulation sentences. Emotion sentences tend to contain words that are experienced in the short term, while \textit{Emotion Regulation} sentences tend to contain features that are stable in the long term.}
\label{fig:flow}
\end{figure}

Traditional personality detection methods either treat text as a whole, relying primarily on direct content analysis~\cite{yang2023psycot,hu2024llm}, or laying particular emphasis on emotion expression~\cite{mohammad2013using,li2021multitask}. However, these approaches often overlook the role of \textbf{Emotion Regulation}~\citep{gross2008emotion}, a key psychological concept related to personality. 
Different from emotion that is usually expressed in a short-term manner, emotion regulation is a stable, long-term status of managing and controlling one's emotional responses, as exemplified in Figure~\ref{fig:flow}. Psychological researches have demonstrated a clear correlation between one's personality and emotion regulation~\cite{baranczuk2019five,petrides2018theory,borges2017role}. 

Inspired by the psychological studies above,  we propose a RAG-based framework named \methodname{} for automatic detection of personality, leveraging both emotion and emotion regulation as guidance. 
To guide the LLMs in personality detection, we first construct a reference library composing of a large number of text-personality pairs. Reference samples most similar to the input text are retrieved to facilitate few-shot learning. To be specific, we categorize each sentence of the input text into emotion sentences and emotion regulation sentences, encode them separately, and then combine the two vectors for effective retrieval of the most similar samples from the reference library. For each retrieved sample, we further include corresponding chain-of-thoughts (CoT) emphasizing emotion and emotion regulation to direct the LLM's attention to these aspects.

For comprehensive evaluation, we test our \methodname{} on both the the Kaggle dataset \footnote{https://www.kaggle.com/datasnaek/mbti-type} for MBTI~\cite{myers1991introduction} detection and the Essays dataset \cite{pennebaker1999linguistic} for the Big Five personality~\cite{goldberg1990alternative} detection. The experimental results show that our method significantly improves the few-shot performance of GPT-3.5 in personality detection tasks, outperforming previous SOTA by 15.5/4.3 on average F1 on the two datasets. Further ablation and analysis consolidates the effectiveness of emotion regulation in personality detection, aligning well with psychological discoveries. To sum up, our contributions are as follows:


\begin{itemize}[leftmargin=*]
    \item To our best knowledge, we are the first to incorporate psychological knowledge of emotion regulation for automatic personality detection.
    \item We propose \methodname{}, a RAG-based framework that combines \textit{Emotion} and \textit{Emotion Regulation} to improve personality detection. Comprehensive experiments on two benchmarks show that our \methodname{} outperforms all the strong baselines by a large margin.
    \item We have conducted in-depth analyses to confirm the effectiveness of  \methodname{} from various aspects, as well as the efficacy of emotion regulation for personality detection.
    
\end{itemize}

\section{Related Work}

\textbf{Personality Detection}\quad
In the early development of personality detection, \citet{francis1993linguistic} introduced the Linguistic Inquiry and Word Count (LIWC), pioneering the use of psycholinguistic features for personality analysis through texts. This tool became foundational for feature engineering in subsequent studies, such as those by \citet{pennebaker1999linguistic} and \citet{argamon2005lexical}, which focused on linguistic styles and lexical predictors of personality traits, achieving moderate accuracies in detecting traits like extraversion and neuroticism.

With the emergence of neural networks, research expanded significantly. Techniques like CNNs and LSTMs enhanced personality prediction from social media \cite{tandera2017personality, xue2018deep}. The introduction of BERT advanced the field further, with \citet{gjurkovic2020pandora} showing its effectiveness in analyzing personality and demographics on Reddit without extensive feature engineering.

Recent studies have explored multi-task and multimodal approaches to personality detection. \citet{sang2022mbti} used movie scripts to predict MBTI types of fictional characters, showing the potential of diverse data integration. \citet{li2021multitask} employed multitask learning to detect emotions and personality traits simultaneously, demonstrating the efficiency of shared representations.

Current research explores large language models (LLMs) for personality detection, as shown by \citet{yang2023psycot} and \citet{hu2024llm}, indicating a shift towards inferring personality traits directly from text with minimal reliance on traditional feature engineering and adopting more holistic, context-aware methodologies.

\noindent \textbf{Emotion and Emotion Regulation}\quad
The relationship between emotion and personality is well-studied in psychology and NLP. Psychological theories \cite{keltner1996facial,davidson2001toward,reisenzein2009personality} link personality traits with emotional experiences, lading the foundation for inferring personality from emotions. \citet{mohammad2013using} showed fine-grained emotions enhance personality detection, \citet{rangra2023emotional} demonstrated the effectiveness of emotional features in speech, and \citet{li2021multitask} found that multitask learning improves prediction accuracy.

\begin{figure*}[t]
\centering
\includegraphics[scale=0.37]{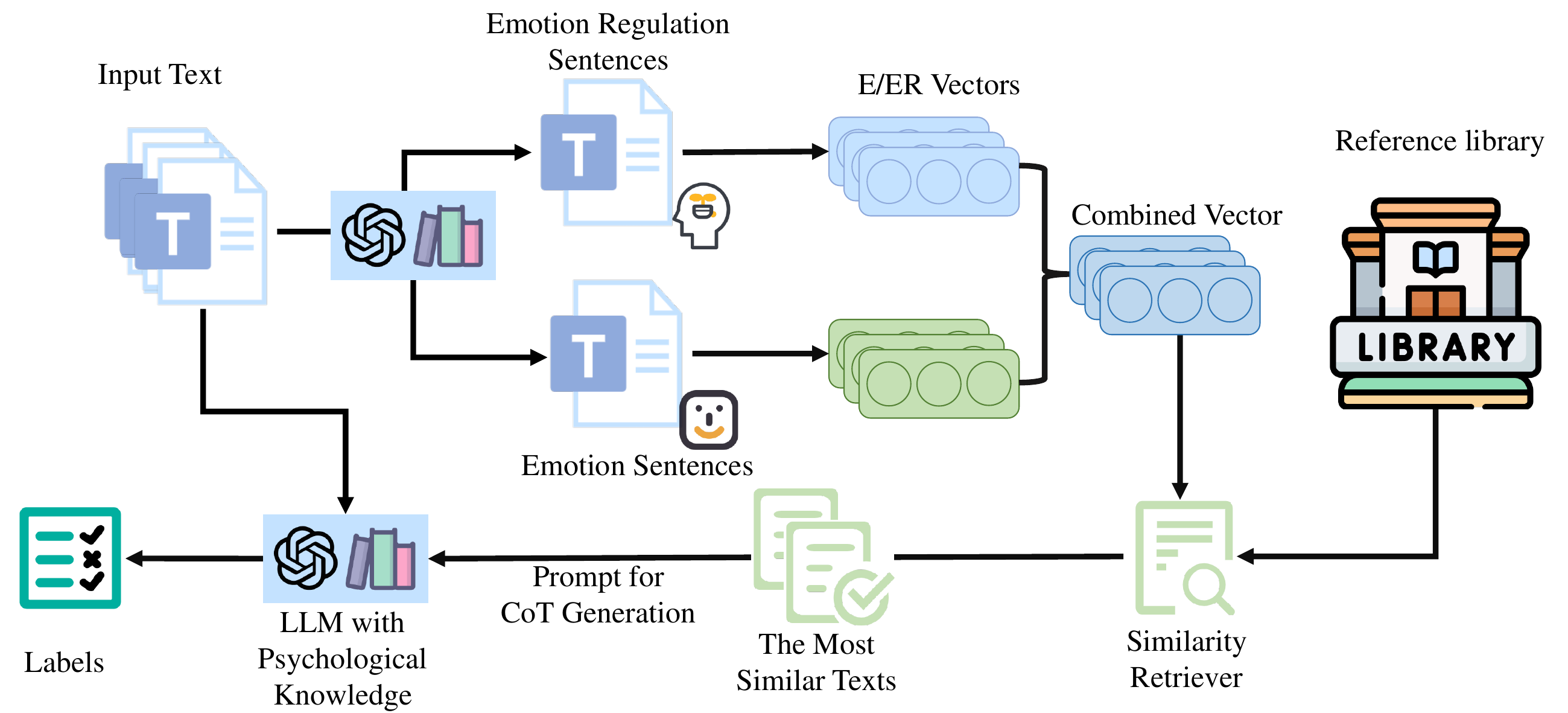}
\caption{An overall framework of \methodname{}. The sentences in input text is categorized into Emotion Sentences and Emotion Regulation Sentences, and then are vectorized and proportionally combined. Using the new vectors, we retrieve two examples and generate their corresponding CoT processes. These examples, along with the input text, are then fed into the LLM with psychological knowledge to obtain the final prediction.}
\label{fig:model}
\end{figure*}

The connection between emotion regulation and personality has also been explored in psychology. Emotion regulation is significantly associated with and can influence personality \cite{baranczuk2019five}. Individuals with strong emotion regulation skills show personality traits assosiated with higher job satisfaction and better stress management \cite{petrides2018theory}. \citet{borges2017role} found emotion regulation variables predict specific personality dimensions. However, to the best of our knowledge, no NLP-based personality detection methods utilize emotion regulation for prediction.



\section{Task Formulation}

In this paper, we focus on the personality detection task, which aims to predict an individual's personality traits from text.
Each text to be detected, $X$, is made up of $\{x_1, x_2, ..., x_n\}$, where each $x_i$ is a sentence.
The goal of personality detection is to map $X$ to a multidimensional label \(y\). 



\section{Method}

In this section, we introduce our \methodname{} framework, as illustrated in Figure~\ref{fig:model}.
First, we construct a reference library, providing text-label pairs to serve as examples for the personality detection model (\S~\ref{sec:refer}).
Then, we utilize psychological knowledge to categorize each sentence in text into Emotion Sentences (ES) and Emotion Regulation Sentences (ERS) (\S~\ref{sec:split}). After that, we retrieve the examples through the combination of ES and ERS (\S~\ref{sec:retrieve}). In inference phase, we utilize examples from reference library for personality detection (\S~\ref{sec:pre}). And the whole method is shown as Algorithm~\ref{algorithm}.

\subsection{Reference Library Construction}
\label{sec:refer}
As the model used for personality detection lacks specialized knowledge in psychology, we employ the Retrieval-Augmented Generation (RAG) method to retrieve and inject relevant examples from the reference library.
We first constructed a reference library, represented as $C = \{(CX_i, y_i)\}_{i=1}^N$, where $CX_i$ and $y_i$ represents the reference input text and personality label of the $i\rm \mbox{-}th$ instance, and $N$ is the size of the reference library.
In our method, we use the training set of the corresponding task as the reference library.

\subsection{Sentence Categorization}
\label{sec:split}
When performing personality detection, emotion and emotion regulation have different characteristics: emotion is expressed in a short-term manner, while emotion regulation is a long-term skill for managing and controlling one's emotional responses. Therefore, we need to handle them separately. 
For the input text $X$, we use a prompt-based approach to have the large language model (LLM) categorize its sentences into two parts. 
We define the classification criteria and input them along with text $X$ into the model, which can then labels each sentence $x$ to perform sentence classification. In this way, the sentences in input text $X$ is categorized into two parts: Emotion Sentences $(X_{e} = \{x_{e1}, x_{e2}, ..., x_{en}\})$ and Emotion Regulation Sentences ($ X_{r} = \{x_{r1}, x_{r2}, ..., x_{rn}\})$. Referring to concepts from psychology~\cite{gross2008emotion}, the classification criteria are defined as follows:


\noindent \textbf{Emotion Sentences}: The feelings in the sentence is dominated by emotion, it should be an obvious reaction to a recent event and not indicative of a deeper, long-standing trait or belief.


\noindent \textbf{Emotion Regulation Sentences}: The feelings in the sentence is dominated by emotion regulation, it should reflect the author's enduring traits rather than immediate circumstances.

For details of the prompt used to accomplish sentence categorization. please refer to Appendix~\ref{sec:appseg}

\begin{algorithm}[t]
\small
\setstretch{1.2}
\caption{\methodname{} Method}\label{algorithm}
\KwIn{Hyperparameter: $\alpha$, \rm{LLM}${\rm{: LLM}}\left( \cdot \right)$, Author's text: X, Reference Library: D, EER split Prompt: $I_p$, $I_m$, Prediction Prompt: PMT
}
\KwOut{The inferred personality trait: y}
$X_{e,r} \leftarrow LLM ({X,I_{e,r}})$\;
$V_{xe,xr} \leftarrow vectorize(X_{e,r})$\;
$V_x \leftarrow V_{xe} + V_{xr}$\;
$Sim \leftarrow [null]$\;
\For{each text $d$ in $D$}{
    $d_e,d_r \leftarrow \rm{LLM}\left( d, I_e,I_r\right)$\;
    $V_{de,dr} \leftarrow vectorize(d_e,d_r)$\;
    $V_d \leftarrow \alpha V_{de} + (1-\alpha) V_{dr}$\;
    $sim \leftarrow 1-\cos(V_x, V_d)$\;
    append $sim$ to $Sim$\;
}
$\{t_{1,2}\} \leftarrow argsort_{t\in D} \ Sim[-2:]$
$Egs \leftarrow \{text_{t_1,t_2}, CoT_{t_1,t_2}\}$\;
$y \leftarrow LLM(PMT, Egs, X)$\;
\Return{y}\;
\end{algorithm}

\begin{figure}[t]
\centering
\includegraphics[scale=1.0]{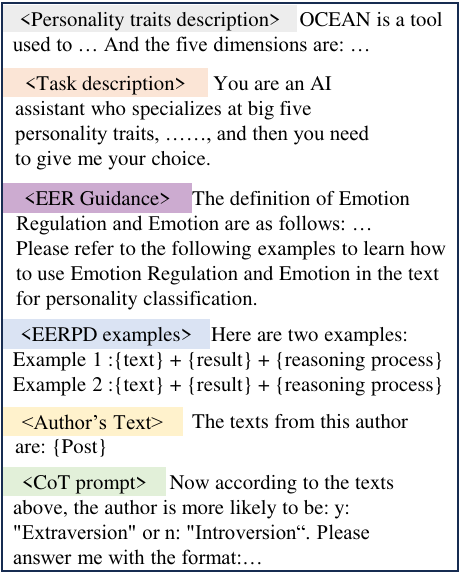}
\caption{An overview of prompt in our method.}
\label{fig:prompt}
\end{figure}


\subsection{Example Retrieval}
\label{sec:retrieve}

People with similar personalities tend to exhibit similar patterns in both emotion and emotion regulation. Therefore, when assessing personality, we retrieve relevant examples from the reference library to assist in detection. To fully utilize both emotion and emotion regulation, we combine them to search for similar examples.


Given the Emotion Sentences \(X_e\) and Emotion Regulation Sentences \(X_r\) of text \(X\), we compute their respective vector representations \(V_{xe}\) and \(V_{xr}\) by the roberta-large model, and then calculate a weighted embedding using a hyperparameter \(\alpha\), such that:
$V_{x} = \alpha V_{xe} + (1-\alpha) V_{xr}$
This hyperparameter \(\alpha\) allows for adjustable emphasis between emotion and emotion regulation influences in the representation and exploring the importance of both for personality detection. Similarly, each text in the reference library is processed to obtain a corresponding composite embedding: 
$V_{d} = \alpha V_{de} + (1-\alpha) V_{dr}$

Then we use 
$Sim(V_{x},V_{d}) = 1 - \cos(V_{x}, V_{d})$ 
to identify the two most analogous texts from the Reference Library. The selected texts, along with their associated CoTs, serve as examples in 2-shot learning for the LLM.

\subsection{Personality Prediction}
\label{sec:pre}
When conducting personality detection, we use psychological knowledge from the MBTI and OCEAN personality dimension models, and generate model prompts using few-shot and CoT learning strategies. In this way, we emphasize emotion and emotion regulation to direct the model's attention to these aspects. The whole prompt is shown in Figure \ref{fig:prompt}, and more details in \ref{sec:apppre}.

\noindent \textbf{Psychological Knowledge.} In the prompt, each personality dimension from MBTI or OCEAN is introduced with a precise psychological definition. For instance, "Extroversion (E) or Introversion (I): indicates whether a person is more inclined to draw energy from the external world or the internal world."

\noindent \textbf{Few-Shot and CoT Learning Strategies.} We retrieve two examples from Reference Library as \S~\ref{sec:retrieve} mentioned, demonstrating how specific personality traits manifest in textual form through emotion and emotion regulation. Also, we leverage LLM to generate CoTs for each example. The two texts along with CoTs are used as 2-shot examples in the prompt. More details about auxiliary CoT generation is shown in Appendix \ref{sec:appgencot}.



\section{Experiments}
\subsection{Datasets}

We conduct an evaluation of \methodname{} using two publicly available datasets: the Kaggle dataset \footnote{https://www.kaggle.com/datasnaek/mbti-type} and the Essays dataset \cite{pennebaker1999linguistic}.

The Kaggle dataset is sourced from PersonalityCafe \footnote{http://personalitycafe.com/forum}, and is an extensive collection of textual data aimed at exploring and predicting personality types based on the Myers-Briggs Type Indicator (MBTI). The personality classification follows the MBTI framework \cite{myers1991introduction}, which segments personality into four dimensions: Introversion/Extraversion (I/E), Sensing/Intuition (S/N), Thinking/Feeling (T/F), and Judging/Perceiving (J/P). The dataset consists of 8,674 entries, each entry representing an individual's text data (each consisting of 45-50 posts) along with their corresponding MBTI type. 

The Essays dataset is a comprehensive collection of text data designed for personality recognition tasks, particularly focusing on the Big Five personality traits \cite{goldberg1990alternative}. Given specific instructions, volunteers wrote freely to express their thoughts within a limited time. 2,468 texts along with each author's Big Five personality traits ( Agreeableness, Conscientiousness, Extraversion, Neuroticism, and Openness) makes up this dataset.


Due to the limited API resources, we randomly selected 10\% samples form each test set to evaluate our \methodname{}.

\begin{table*}[t]
	\renewcommand{\arraystretch}{1.0}
	\centering
	\resizebox{0.85\textwidth}{!}{
		\begin{tabular}{ccccccccccc}
			\toprule
			\multirow{2}{*}{\textbf{Methods}} & \multicolumn{2}{c}{\textbf{I/E}} & \multicolumn{2}{c}{\textbf{S/N}} & \multicolumn{2}{c}{\textbf{T/F}} & \multicolumn{2}{c}{\textbf{J/P}} & \multicolumn{2}{c}{\textbf{Average}} \\ 
			& Acc. & F1 & Acc. & F1 & Acc. & F1 & Acc. & F1 & Acc. & F1 \\
			\hline \hline
			TF-IDF+SVM & 71.00 & 44.94 & 79.50 & 46.38 & 75.00 & 74.25 & 61.50 & 58.59 & 71.75 & 56.04 \\
                Regression & 61.34 & 64.00 & 47.10 & 54.50 & 76.34 & 76.50 & 65.58 & 66.00 & 62.59 & 65.25 \\
                AttRCNN & - & 59.74 & - & 64.08 & - & 78.77 & - & 66.44 & - & 67.25 \\
			TrigNet & 77.80 & 66.64 & 85.00 & 56.45 & 78.70 & 78.32 & 73.30 & 71.74 & 78.70 & 68.29 \\
                DDGCN & 78.10 & 70.26 & 84.40 & 60.66 & 79.30 & 78.91 & 73.30 & 71.73 & 78.78 & 70.39 \\
                BERT & 77.30 & 62.50 & 84.90 & 54.04 & 78.30 & 77.93 & 69.50 & 68.80 & 77.50 & 65.82 \\
			RoBERTa & 77.10 & 61.89 & 86.50 & 57.59 & 79.60 & 78.69 & 70.60 & 70.07 & 78.45 & 67.06 \\
			\midrule
			Zero-shot-CoT & 76.50 & 64.27 & 83.50 & 55.16 & 72.50 & 71.99 & 57.50 & 53.14 & 72.50 & 61.14 \\
                Two-shot-CoT & \underline{85.93} & \underline{85.41} & 78.89 & \underline{77.55} & \underline{87.44} & \underline{86.77} & \underline{69.35} & \underline{70.36} & \underline{80.40} & \underline{80.02} \\
                TAE & - & 70.90 & - & 66.21 & - & 81.17 & - & 70.20 & - & 72.07 \\
			PsyCoT & 79.00 & 66.56 & \underline{85.00} & 61.70 & 75.00 & 74.80 & 57.00 & 57.83 & 74.00 & 65.22 \\
                \methodname{}(our) & \textbf{87.10} & \textbf{86.63} & \textbf{91.01} & \textbf{90.59} & \textbf{89.17} & \textbf{89.15} & \textbf{81.34} & \textbf{82.12} & \textbf{87.15} & \textbf{87.12} \\
			\bottomrule
	\end{tabular}}
        \caption{Overall results of \methodname{} and baselines on the Kaggle dataset.}
	\label{tab:kaggle}
\end{table*}

\begin{table*}[t]
	\renewcommand{\arraystretch}{1.0}
	\centering
	\resizebox{1\textwidth}{!}{
		\begin{tabular}{ccccccccccccc}
			\toprule
			\multirow{2}{*}{\textbf{Methods}} & \multicolumn{2}{c}{\textbf{AGR}} & \multicolumn{2}{c}{\textbf{CON}} & \multicolumn{2}{c}{\textbf{EXT}} & \multicolumn{2}{c}{\textbf{NEU}} & \multicolumn{2}{c}{\textbf{OPN}} & \multicolumn{2}{c}{\textbf{Average}} \\ 
			& Acc. & F1 & Acc. & F1 & Acc. & F1 & Acc. & F1 & Acc. & F1 & Acc. & F1 \\
			\hline \hline
			LIWC+SVM & 51.78 & 47.50 & 51.99 & 52.00 & 51.22 & 49.20 & 51.09 & 50.90 & 54.05 & 52.40 & 52.03 & 50.40 \\
            Regression & 50.96 & 51.01 & 54.65 & 54.66 & 55.06 & 55.06 & 57.08 & \underline{57.09} & 59.51 & 59.51 & 55.45 & 55.47 \\
			W2V+CNN & - & 46.16 & - & 52.11 & - & 39.40 & - & \textbf{58.14} & - & 59.80 & - & 51.12 \\
			BERT & 56.84 & 54.72 & 57.57 & 56.41 & 58.54 & 58.42 & 56.60 & 56.36 & 60.00 & 59.76 & 57.91 & 57.13 \\
			RoBERTa & 59.03 & 57.62 & 57.81 & 56.72 & 57.98 & 57.20 & \underline{56.93} & 56.80 & 60.16 & \underline{59.88} & 58.38 & 57.64 \\
			\midrule
			Zero-shot-CoT & 58.94 & 58.09 & 55.14 & 42.49 & 57.55 & 55.63 & \textbf{57.49} & 54.63 & 58.78 & 54.40 & 57.58 & 53.05 \\
                Two-shot-CoT & 55.06 & 57.27 & 59.51 & \underline{59.63} & 52.63 & 52.84 & 53.85 & 53.64 & 57.09 & 57.74 & 55.63 & 56.22 \\
			PsyCoT & \underline{61.13} & \underline{61.13} & \underline{59.92} & 57.41 & \underline{59.76} & \underline{59.74} & 56.68 & 56.58 & \underline{60.73} & 57.30 & \underline{59.64} & \underline{58.43} \\
                \methodname{}(our) & \textbf{64.98} & \textbf{65.01} & \textbf{68.00} & \textbf{68.64} & \textbf{62.01} & \textbf{63.02} & 56.00 & 56.00 & \textbf{61.02} & \textbf{60.93} & \textbf{62.40} & \textbf{62.72} \\
			\bottomrule
	\end{tabular}}
        \caption{Overall results of \methodname{} and baselines on the Essays dataset. We use Accuracy(\%) and Macro-F1(\%) as metrics. Best results are listed in bold and the second best results are shown with underline.}
	\label{tab:essays}
\end{table*}

\subsection{Baselines}
In our experiments, we adopt the following previous methods as baselines.

\noindent \textbf{Statistical Learning}: These methods aim to enhance sentiment classification accuracy through statistical learning methods. \citet{tighe2016personality} uses SVM with LIWC \cite{pennebaker2001linguistic} and linguistic cognitive analysis. \cite{cui2017survey} uses SVM with TF-IDF for feature extraction. \cite{park2015automatic} uses ridge regression to conduct regression modeling between the language features and users' Big Five personality traits. 

\noindent \textbf{Neural Network Models}: These methods leverage neural network architectures to enhance personality detection. W2V+CNN \cite{rahman2019personality} is a non-pretrained CNN model \cite{chen2015convolutional} combined with the word2vec algorithm for context representation. AttRCNN \cite{attRCNN} uses a hierarchical structure with a variant of Inception \cite{szegedy2017inception} to encode each post. DDGCN \cite{yang2022orders} employs a domain-adapted BERT to encode each post and a dynamic deep graph network to aggregate posts non-sequentially. Small language models like BERT \cite{devlin2019bert} and RoBERTa \cite{liu2019roberta} are fine-tuned on "\texttt{bert-base-cased}" and "\texttt{roberta-base}" backbones, encoding the context for Essays and combining post representations using mean pooling for Kaggles.

\noindent \textbf{Large Language Models}: These methods either use LLMs directly or incorporate them as a significant component of the model architecture. \citet{kojima2022large} inserts a reasoning step with \textit{"Let's think step by step"}, and is adopted as the Zero-shot-CoT baseline in this work. TAE \cite{hu2024llm} improves small model performance in personality detection using text augmentations from LLMs and contrastive learning. PsyCoT \cite{yang2023psycot} uses psychological questionnaires as a CoT process, leveraging LLM to perform multi-turn dialogue ratings. We also build a Two-shot CoT prompt as a reference baseline for our \methodname{}, consisting of two randomly selected examples.

\subsection{Implementation Details}
Due to baseline research and economic considerations, we request the GPT-3.5 API (\texttt{gpt-3.5-turbo-16k-0613}) to obtain results, which is currently the most popular and forms the foundation of ChatGPT. For Zero-shot-CoT, Two-shot-CoT and our \methodname{} methods, we set the temperature to 0 to get a reliable rather than innovative output. For the PsyCoT and TAE method, we adopt the results from \citet{yang2023psycot} and \citet{hu2024llm}. For the AttRCNN and W2V+CNN, we adopt the results from \citet{hu2024llm}, setting the learning rate for the pre-trained post encoder to 1e-5, and for other parameters 1e-3. For the other fine-tuning based methods, we adopt the baseline results directly from \citet{yang2023psycot}, where the learning rate was set to 2e-5 and the test performance was evaluated by averaging the results of five runs. The evaluation metrics employed in our study include Accuracy and Macro-F1 score.

\subsection{Overall Results}
The overall results of \methodname{} and several baselines on Kaggle are listed in Table \ref{tab:kaggle}, and on Essays are listed in Table \ref{tab:essays}. The small model baselines can divided into three types: statistical learning models (LIWC+SVM, TF-IDF+SVM, Regression), convolution models (W2V+CNN, AttRCNN, TrigNet, DDGCN), and small language models (BERT and RoBERTa). The baselines involved in large language models are: Zero-shot-CoT, Two-shot-CoT, TAE and PsyCoT. 

Several key points emerge from these results: 

\textbf{First}, \methodname{} outperforms the baselines on almost all the personality traits, surpassing the fine-tuned models and other prompt-based methods. Specifically, \methodname{} enhances standard Two-shot-CoT prompting with an average increase of \textbf{6.75/7.10} points in Accuracy and Macro-F1 on Kaggle, \textbf{6.77/6.50} in Accuracy and Macro-F1 on Essays. 

\textbf{Second}, \methodname{} performs worse than other methods on the Neuroticism trait. Further analysis reveals that this discrepancy may be due to the low correlation between language-based assessments and self-report questionnaires for Neuroticism. As shown in \citet{park2015automatic}, Neuroticism has the lowest correlation coefficients with self-report questionnaires, indicating that language models struggle to accurately capture and predict this trait, leading to lower prediction accuracy. 

\textbf{Third}, although includes two examples with CoT, Two-shot-CoT baseline does not consistently improve the performance of PsyCoT or even Zero-shot-CoT on Essays. Our investigation shows that the examples in Two-shot-CoT are sometimes unhelpful or even negative for detection in Essays, with high conflicts with the sample to be tested. For instance, if both given examples have high Agreeableness, even if the author criticizes their roommate in the test sample, the reasoning still considers it to be high agreeable.

\begin{table}[t]
\centering
\resizebox{1\columnwidth}{!}{
\begin{tabular}{l|ccccc}
\hline
\multirow{2}{*}{\textbf{Methods}} & \multicolumn{5}{c}{\textbf{Kaggle}}                                                                                                   \\ \cline{2-6} 
                                  & \textit{\textbf{I/E}} & \textit{\textbf{S/N}} & \textit{\textbf{T/F}} & \multicolumn{1}{c|}{\textit{\textbf{P/J}}} & \textbf{Average} \\ \hline

\methodname{}\textsubscript{r/w 0-shot}        & 76.50 & 83.50 & 72.50 & \multicolumn{1}{c|}{57.50} & 72.50 \\
\methodname{}\textsubscript{r/w 2-shot} & 85.93 & 78.89 & 87.44 & \multicolumn{1}{c|}{69.35} & 80.40 \\ \hline

\methodname{}\textsubscript{w/o E}             & 71.24 & 66.34 & 80.61 & \multicolumn{1}{c|}{67.43} & 71.40 \\
\methodname{}\textsubscript{w/o ER}            & 70.14 & 65.29 & 80.03 & \multicolumn{1}{c|}{69.55} & 71.25 \\ \hline
\methodname{}                                  & 87.10 & 91.01 & 89.17 & \multicolumn{1}{c|}{81.34} & 87.15 \\ \hline
\end{tabular}}
\caption{Results of ablation study on Accuracy on the Kaggle dataset.}
\label{tab:ablation}
\end{table}

\subsection{Ablation Study} \label{sec:abla}

To verify the importance of each component in our \methodname{}, we conduct an ablation study on 100 samples randomly selected from each of the Kaggle test dataset and Essays test dataset.

\paragraph{Emotion and Emotion Regulation.} We first analyze the contributions of Emotion and Emotion Regulation for example retrieval. $\methodname{}{w/o\ E}$ refers to the condition where $\alpha=1.0$, meaning examples are retrieved only based on Emotion Regulation Vectors. $\methodname{}{w/o\ ER}$ refers to the condition where $\alpha=0.0$, meaning examples are retrieved only based on Emotion Vectors. As shown in Table \ref{tab:ablation}, we use three groups for comparison: our overall method shows significant improvements compared to the 0-shot and 2-shot baselines; compared to baselines utilizing only Emotion Regulation ($\methodname{}{w/o\ E}$) or only Emotion ($\methodname{}{w/o\ ER}$), it is demonstrated that both components contribute to improvements. Our combined method outperforms both $\methodname{}{w/o\ E}$ and $\methodname{}{w/o\ ER}$

\begin{figure}[t]
\centering
\includegraphics[scale=0.5]{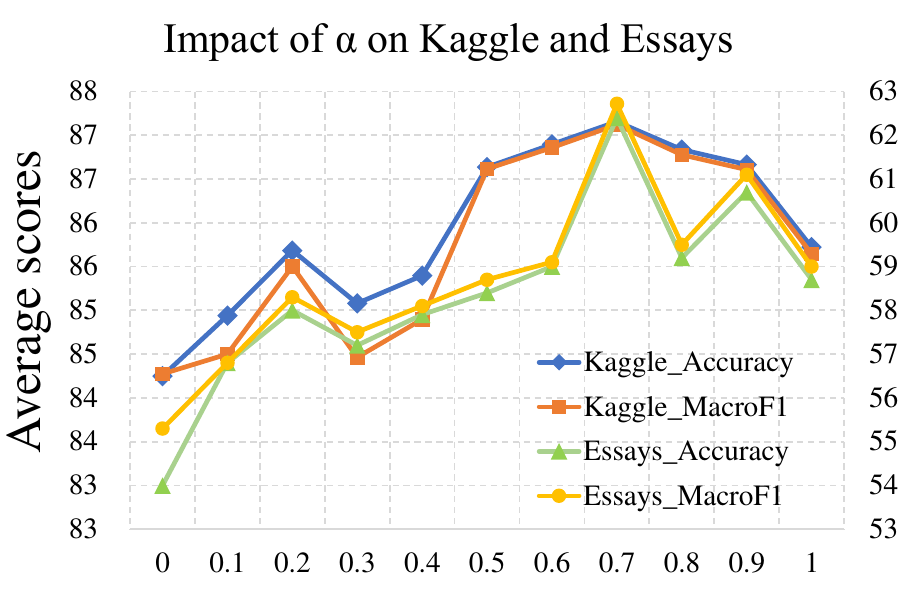}
\caption{Impact of Alpha on Kaggle and Essays.}
\label{fig:alpha}
\end{figure}

\paragraph{Parameter $\alpha$.} We investigate the trade-off parameter $\alpha$ in our \methodname{}, demonstrating the model's sensitivity to $\alpha$ variations and identifying its optimal range. The results, shown in Figure \ref{fig:alpha}, illustrate performance variations with $\alpha$ on the Kaggle and Essays test datasets.

For the two dataset, both accuracy and Macro-F1 score peaked at $\alpha=0.7$, then declined. These findings suggest that $\alpha=0.7$ has the best balance between Emotion Regulation and Emotion, with emotion regulation proving more predictive of personality. Performance drops when $\alpha$ is 0 or 1, highlighting that combining both features is more effective than using a single one. The experimental results also show that $\alpha=0.7$, is also higher than $\alpha=0.5$, which is the vector of the original text directly used for retrieval without combining in weighted proportion. And it proves that our combined method is more efficient than the whole article for retrieval.

\paragraph{Auxiliary CoT.} We evaluate the effect of the auxiliary CoTs generation statistically, using the 100 random samples from Kaggle dataset. And the F1 scores are shown in Table \ref{tab:cot}. Each data point is the mean of 5 trials. T-test analysis demonstrates that the auxiliary CoTs has statiscal significance with p less than 0.05.

\begin{table}[h]
	\renewcommand{\arraystretch}{1.0}
	\centering
        \setlength\tabcolsep{7pt}
		\begin{tabular}{lccccc}
			\toprule
			\textbf{Dataset} & \textbf{E/I} & \textbf{N/S} & \textbf{T/F} & \textbf{P/J} \\
			\hline \hline
                \methodname{} & 90.56 & 91.97 & 92.43 & 81.51\\
			  \ {w/o CoTs} & 78.61 & 83.17 & 86.88 & 79.03 \\
			\bottomrule
	\end{tabular}
        \caption{Results of the ablation study on auxiliary CoTs. \methodname{} with CoTs outperforms w/o baseline statistically.}
	\label{tab:cot}
\end{table}

\section{Analysis}
\subsection{Different Base Model}

\begin{figure}[t]
\centering
\includegraphics[scale=0.3]{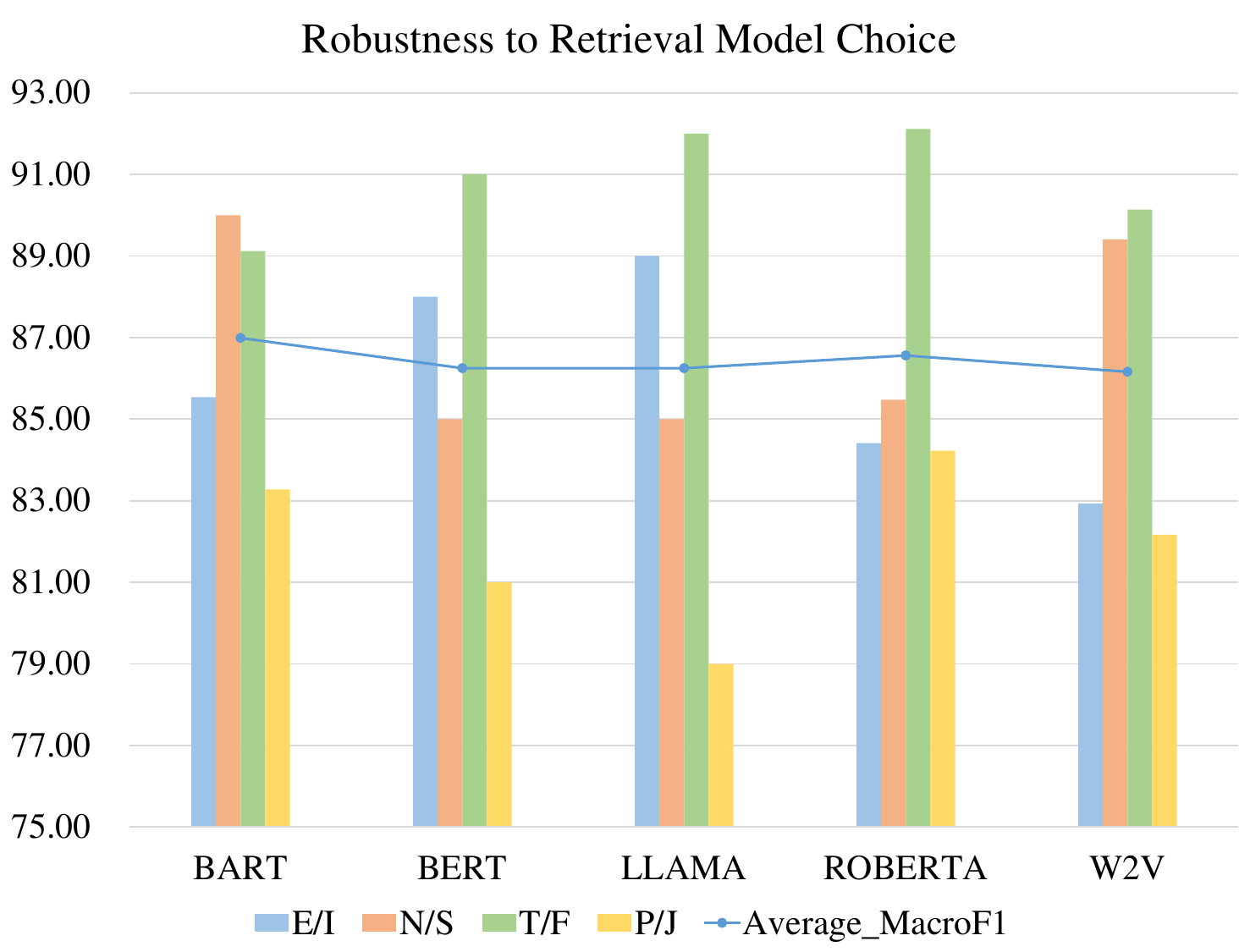}
\caption{The Performance of Different Retrieval Models on Kaggle dataset.}
\label{fig:robustness}
\end{figure}
\begin{figure}[t]
\centering
\includegraphics[scale=0.8]{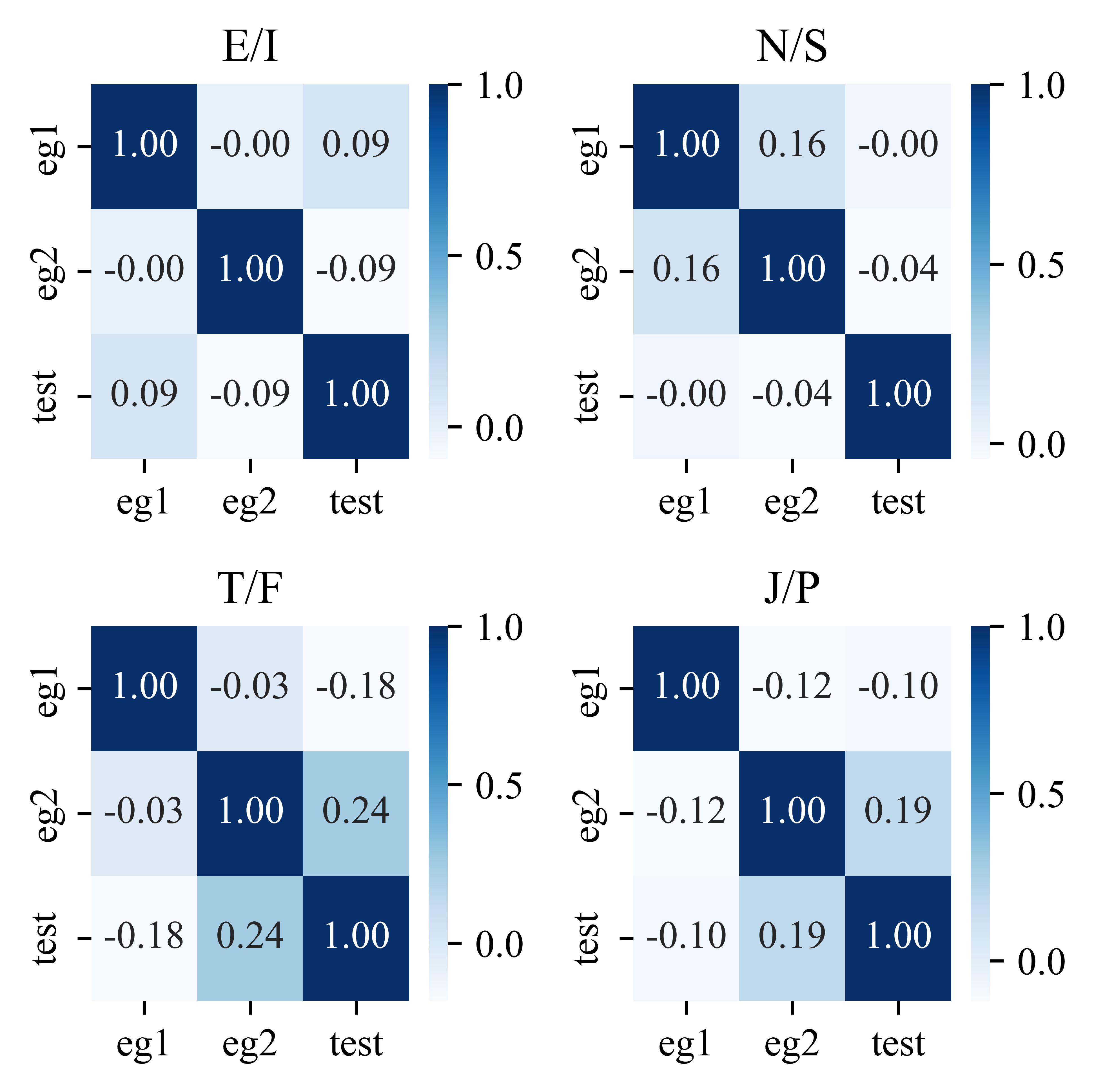}
\caption{Correlation Analysis on Example Selection.}
\label{fig:corr}
\end{figure}

To evaluate the robustness of our \methodname{} method across different model architectures, we conducted experiments using various popular language models, including BART \cite{lewis2019bart}, BERT \cite{devlin2018bert}, LLAMA \cite{touvron2023llama}, RoBERTa \cite{liu2019roberta}, and W2V \cite{le2014distributed}. The evaluation dataset consists of 100 samples selected from the Kaggle test dataset.

The results presented in Figure \ref{fig:robustness}, indicate that \methodname{} maintains consistent performance across different model architectures. This flexibility allows for broader applicability in various settings.

\subsection{Correlation Analysis on Example Selection}
To evaluate our example selection, we conducted a correlation test between the selected examples and the test set examples. As shown in Figure \ref{fig:corr}, the result reveals no significant correlation, confirming that our selection method does not leak test set answers to the model. Instead, it identifies examples with similar reasoning patterns. This demonstrates that our method effectively teaches the model relevant reasoning techniques, ensuring that it learns to generalize rather than memorize specific answers. Thus, our approach enhances the model's ability to perform accurate label predictions based on learned reasoning strategies.

\begin{table}[t]
	\renewcommand{\arraystretch}{1.0}
	\centering
        \setlength\tabcolsep{8pt}
		\begin{tabular}{lccccc}
			\toprule
			\textbf{Dataset} & \textbf{E/I} & \textbf{N/S} & \textbf{T/F} & \textbf{P/J} \\ 
			\hline \hline
   
			Order &  84.16 & 86.03 &	91.14 &	80.74 \\
                Random & 85.54 &	89.40 &	89.02 &	79.61 \\
			\bottomrule
	\end{tabular}
        \caption{Results of the study on post orders. The performance of our \methodname{} is not significantly affected.}
	\label{tab:order}
\end{table}

\subsection{Impact of Post Order}

The Kaggle dataset includes a collection of posts for each user. These posts are combined in sequence to form a lengthy document, and each post is an input $X$ waiting for detection. However, as \citet{yang2023psycot} mentioned, researches by \citet{yang2021multi,yang2022orders} have shown that sequential encoding of posts is sensitive to order in fine-tuned models. To determine if \methodname{} are affected by post order, we randomly shuffled the posts and re-evaluated our method using 100 samples. And the F1 results are printed in Table \ref{tab:order}. Each score is the average result after five rounds of experiments. The T-test analysis indicates that there is no statistically significant difference between the sequential test and the random-shuffled test.

\begin{table}[t]
	\renewcommand{\arraystretch}{1.0}
	\centering
        \setlength\tabcolsep{8pt}
		\begin{tabular}{lccccc}
			\toprule
			\textbf{Dataset} & \textbf{E/I} & \textbf{N/S} & \textbf{T/F} & \textbf{P/J} \\
			\hline \hline
   
			Standard &  84.23 & 85.26 & 87.33 & 79.27\\
                \methodname{} & 90.56 & 91.97 & 92.43 & 81.51\\
			\bottomrule
	\end{tabular}
        \caption{Results of the study on post orders. The performance of our \methodname{} significantly outperforms standard Two-Shot-CoT baseline.}
	\label{tab:sta}
\end{table}

\subsection{Statistical Tests}

To statistically evaluate the significance of our approach, we compared the standard Two-shot-CoT baseline with \methodname{} using the 100 random samples from Kaggle dataset, and the F1 scores are shown in Table \ref{tab:sta}. Each data point is the mean of 5 trials. T-test analysis demonstrates that our enhancements has statiscal significance, with p-values of less than 0.05 for each dimension.

\section{Conclusion}

In this paper, we introduced \methodname{}, a novel few-shot personality detection method inspired by psychological concepts. By leveraging emotion regulation and emotion to retrieve few-shot samples, \methodname{} forms a robust Chain of Thought (CoT) that guides large language models in evaluating personality traits. Experiments on two benchmark datasets show that \methodname{} significantly outperforms traditional methods and other novel prompts. This approach uniquely integrates psychological insights to enhance the reasoning abilities of large language models, offering a new perspective for personality detection.

\section*{Limitations}

Due to the resource constraints, we only conduct experiments and analysis about LLMs on GPT-3.5. The extent to which GPT-4 or GPT-4o models can benefit from our \methodname{} remains unknown.

This study primarily focuses on improving the LLM's performance by leveraging the psychological knowledge of Emotion Regulation. How to exploit trainable and tunable models like BERT and LLAMA to further optimize \methodname{} is left for future investigation.

This method carries certain potential risks. Even with well-intentioned use, personality detection may lead to misjudgments, negatively impacting individuals' careers or social relationships.

\section*{Ethics Statement}
This work adheres to the ACL Ethics Policy. We assert that, to the best of our knowledge, our work does not present any ethical issues. We have conducted a thorough review of potential ethical implications in our research and found none.




\appendix

\section{Appendix: Prompt for Prediction}
\label{sec:apppre}
To better understand our method, we provide all the prompt in appendix, and record for the whole prediction is in Figure \ref{fig:preprompt}.

\section{Appendix: Prompt for Sentence Categorization}
\label{sec:appseg}
\begin{figure*}[t]
{\footnotesize\begin{lstlisting}[frame=trBL,breaklines,columns=flexible]
# Record of prediction prompt

MBTI is a tool used to assess a person's psychological preferences and personality types, and there are 16 different types of MBTI, each consisting of four letters representing four dimensions of preference. And the four dimensions are:
Extroversion (E) or introversion (I) : indicates whether a person is more inclined to draw energy from the outside world or the inside world.
Sense (S) or intuition (N) : indicates whether a person is more inclined to focus on concrete facts and details, or abstract concepts and possibilities.
Thinking (T) or emotion (F) : indicates whether a person is more inclined to make decisions using logic and principles, or values and emotions.
Judgment (J) or perception (P) : indicates whether a person is more inclined to a planned and organized lifestyle, or a flexible and random lifestyle.
You are an AI assistant who specializes at MBTI personality traits. I will give you texts from the same author, and then I will ask you the author's MBTI type, and then you need to give me your choice. 

The definition of Emotion Regulation and Emotion are as follows: 
1.Emotion Sentences: These sentences should be clearly linked to immediate, temporary feelings that arise from specific, recent incidents or current situations. The key is that the emotion should be an obvious reaction to a recent event and not indicative of a deeper, long-standing trait or belief.
2. Emotion Regulation Sentence: These sentences must consistently reflect the speaker's enduring traits. They should not be influenced by immediate circumstances but rather indicate a persistent and characteristic ability of controling emotion.

Please refer to the following examples to learn how to use Emotion Regulation and Emotion in the text for personality classification.

I will give you 45~50 posts from the same user, divided by |||. Please use MBTI personality analysis to help me analyze what the user's MBTI is most likely to be. 
Here are two examples:
---
Example 1:
The posts of this user are: +cot.iloc[e1]['posts']+
Result: """+cot.iloc[e1]['type']+"""
Process:"""+cot.iloc[e1]['cot']+"""
---
Example 2
The posts of this user are: """+cot.iloc[e2]['posts']+"""
Result: """+cot.iloc[e2]['type']+"""
Process:"""+cot.iloc[e2]['cot']+"""
---
Now, analysis the user's MBTI type with your reasoning process.
It should be noted that when the user's first dimension result is E, the user's fourth dimension result is more likely to be P.
The user's post reads as follows: 
"""+post+"""
Your response should follow the following format: 
Result: {The four letters represent the type of mbti you guessed}
Process: {your reasoning process}.

\end{lstlisting}}
\caption{A complete record of prediction}
\label{fig:preprompt}
\end{figure*}

To better understand our method, we provide all the prompt in appendix, and record for EER Text Split prompt is in Figure \ref{fig:splitprompt}.

\begin{figure*}[t]
{\footnotesize\begin{lstlisting}[frame=trBL,breaklines,columns=flexible]
# Record of EER Text Split

I am a sentence sentiment adjudicator specialized in distinguishing the sources of emotions in sentences - whether they stem from the speaker's current mood or their inherent personality. Your task is to assist me by examining the text and discerning the dominant influence in each sentence, based on these highly refined definitions:

1. Emotion Sentences: These sentences should be clearly linked to immediate, temporary feelings that arise from specific, recent incidents or current situations. The key is that the emotion should be an obvious reaction to a recent event and not indicative of a deeper, long-standing trait or belief.
• Highly Refined Definition: Look for signs that the emotion is an immediate response to a particular event, is temporary, and doesn't reflect an ongoing pattern of thoughts or behaviors.
• Examples and Analysis:
  - "Sex can be boring if it's in the same position often. For example me and my girlfriend are currently in an environment where we have to creatively use cowgirl and missionary. There isn't enough..." - Emotion, as it describes a current, specific situation causing temporary boredom.
  - "I'm thrilled about the concert tonight!" - Emotion, as the excitement is tied to a specific, imminent event (the concert).
  - "I am anxious because of the upcoming exam." - Emotion, since the anxiety is a temporary response to a specific future event (the exam).
  - "I am angry with my friend for something they did last week." - Emotion, because the anger is a reaction to a specific, recent event (the friend's action last week).

2. Emotion Regulation Sentence: These sentences must consistently reflect the speaker's enduring traits. They should not be influenced by immediate circumstances but rather indicate a persistent and characteristic ability of controling emotion.
• Highly Refined Definition: Determine if the expression is reflective of a longstanding personality trait for emotion control, consistent over time and not a reaction to a specific, recent circumstance.
• Examples and Analysis:
  - "I'm finding the lack of me in these posts very alarming." - Emotion Regulation, as it reflects a long-term concern about self-representation rather than an immediate emotional reaction.
  - "Giving new meaning to 'Game' theory." - Emotion Regulation, as it expresses a general viewpoint on a concept, not about temporary feelings.
  - "Hello *ENTP Grin* That's all it takes. Than we converse and they do most of the flirting while I acknowledge their presence and return their words with smooth wordplay and more cheeky grins." - Emotion Regulation, as it describes a consistent behavior pattern rather than a reaction to a specific event.
  - "Real IQ test I score 127. Internet IQ tests are funny. I score 140s or higher. Now, like the former responses of this thread I will mention that I don't believe in the IQ test. Before you banish..." - Emotion Regulation, as it reflects a long-standing skepticism towards IQ tests, not an immediate emotional reaction.

Special Case: Any sentences containing only a URL should be classified as 'Emotion Regulation'.
  - "http://www.youtube.com/watch?v=4V2uYORhQOk" - Emotion Regulation, because it is a pure URL.
  - "http://playeressence.com/wp-content/uploads/2013/08/RED-red-the-pokemon-master-32560474-450-338.jpg" - Emotion Regulation, as it is a URL.

Ambiguous Examples and Detailed Analysis:
1. "The last thing my INFJ friend posted on his facebook before committing suicide the next day. Rest in peace~" - Emotion. Although it mentions an INFJ personality type, the focus is on the immediate emotional reaction to the friend's recent suicide.
2. "I often find myself reflecting deeply on my experiences." - Emotion Regulation. This indicates a consistent trait of introspection, not linked to a specific, recent event.

For each sentence provided, carefully determine whether it primarily reflects 'Emoiton' or 'Emotion Regulation', based on these highly refined criteria. List each sentence and categorize it as either 'Emotion' or 'Emotion Regulation'.

The texts from this author are: """ + post + """.

Respond in the following format without any reason or explain:
0.  [Emotion/Emotion Regulation]
1.  [Emotion/Emotion Regulation]
2.  [Emotion/Emotion Regulation]

Focus meticulously on these criteria to maximize the accuracy and consistency of classification.

\end{lstlisting}}
\caption{A complete record that \methodname{} is applied to split the text into E or ER part.}
\label{fig:splitprompt}
\end{figure*}

\section{Appendix: Prompt for Generation of auxiliary CoT}
\label{sec:appgencot}
To better understand our method, we provide all the prompt in appendix, and record for generating the CoT in Reference Library is in Figure \ref{fig:gencot}.

\begin{figure*}[t]
{\footnotesize\begin{lstlisting}[frame=trBL,breaklines,columns=flexible]
# Record of CoT Generation

Suppose you are a psychologist with a keen interest in personality types and online behavior. You know that MBTI is a tool used to assess a person's psychological preferences and personality types, and there are 16 different types of MBTI, each consisting of four letters representing four dimensions of preference. And the four dimensions are:

Extroversion (E) or introversion (I) : indicates whether a person is more inclined to draw energy from the outside world or the inside world.
Sense (S) or intuition (N) : indicates whether a person is more inclined to focus on concrete facts and details, or abstract concepts and possibilities.
Thinking (T) or emotion (F) : indicates whether a person is more inclined to make decisions using logic and principles, or values and emotions.
Judgment (J) or perception (P) : indicates whether a person is more inclined to a planned and organized lifestyle, or a flexible and random lifestyle.

I will give you 45~50 posts from the same user, divided by |||. Please use MBTI personality analysis to help me analyze what the user's MBTI is most likely to be. I will give you 45~50 posts from the same user, divided by |||, and the MBTI type of the user. Please use MBTI personality analysis to help me analyze why the user is this MBTI type. 
Here is an example:
---
Example:
The posts of this user are:  'Wow, thank you for this thread! Physical vs. metaphysical is a great topic! I find that I am very much the same way your are. How can I put it....I have my days. :) The more I develop my xSxJ, the...|||my room. I like to be in my bad, next to my books, with my fan on and laptop nearby.|||I wouldn't say that I can read souls - but I can see potential. I can sense sadness, happiness, uneasiness, etc. I can tell when someone is not happy where they are and with what they are doing with...|||thank you for being so polite! :)|||I find eye contact is key. I acknowledge their existence and importance by maintaining eye contact with them throughout the conversation. Not by staring in their eyes in a creeper way, but by making...|||As an INFJ male I can somewhat relate to your post. A very close lady friend of mine and I were like this for years! I had always liked her and could read her fairly well. I knew when she needed...'

Result: INFJ

Process: Based on the posts you provided, I would guess that the user is an INFJ personality type. INFJs are known as the advocates, who are quiet and mystical, yet very inspiring and tireless idealists. They are often deeply spiritual, compassionate, and intuitive. They value harmony, authenticity, and personal growth. They can also be very sensitive, private, and perfectionistic.
Some clues that suggest the user is an INFJ are:
First of all, I think the user is an introvert (I). The user prefers to spend time alone in his room with books and laptop, rather than socializing with many people. He also seem to be more focused on his inner world of thoughts and feelings, rather than the outer world of events and actions.
Secondly, I think the user is an intuitive (N). He is interested in abstract concepts and possibilities, such as physical vs. metaphysical. He can see the potential in people and situations, and he is not limited by the facts and details. He also has a wide range of knowledge and interests, and he is constantly learning and innovating.
Thirdly, I think the user is a feeler (F). He makes decisions based on his values and emotions, rather than logic and principles. He can sense the emotions of others and empathize with them. He is polite and respectful, and he values harmony and cooperation.
Lastly, I think the user is a judger (J). He prefers a planned and organized lifestyle, rather than a flexible and random one. He has a clear sense of direction and purpose, and he likes to achieve his goals. He also have a strong xSxJ side, which means he can use his sensing function to deal with reality and details when necessary.
Therefore, based on my analysis, I think the user's MBTI type is INFJ. INFJs are known as the advocates or the counselors. They are idealistic, creative, compassionate, and insightful. They have a vision of how to make the world a better place, and they use their intuition and feeling to inspire and motivate others. They are also loyal, dedicated, and supportive of their friends and loved ones.
---
Now, you should generate the {Process}, according to the MBTI type and the posts given to you.
The user's MBTI type is: """+type+""", and the user's posts are:"""+post+""".
Your response should follow the following format: 
Process: {your reasoning process}.

"""

\end{lstlisting}}
\caption{A complete record for the Generation of auxiliary CoT in Reference Library.}
\label{fig:gencot}
\end{figure*}

\end{document}